\documentclass[conference]{IEEEtran}

\IEEEoverridecommandlockouts                              

\title{Deep Reinforcement Learning using Genetic Algorithm for Parameter Optimization}

\author{Adarsh Sehgal, Hung Manh La, Sushil J. Louis, Hai Nguyen
\thanks{Adarsh Sehgal, Hai Nguyen and Dr. Hung La are with the Advanced Robotics and Automation
(ARA) Laboratory. Dr. Sushil Louis is professor of the Department of Computer Science and Engineering, University of Nevada, Reno, NV 89557, USA. Corresponding author: Hung La, email: {\tt\small hla@unr.edu}}
\thanks{This material is based upon work supported by the National Aeronautics and Space Administration (NASA) Grant No. NNX15AI02H issued through the NVSGC-RI program under sub-award No. 19-21, and the RID program under sub-award No. 19-29, and the NVSGC-CD program under sub-award No. 18-54. This work is also partially supported by the Office of Naval Research under Grant N00014-17-1-2558.}
}

\usepackage{amsmath}
\usepackage{algorithm}
\usepackage{algpseudocode}
\usepackage{graphicx}
\usepackage{subcaption}
\usepackage{lipsum}
\usepackage{multicol}
\usepackage{cite}
\graphicspath{{images/}}

\begin{document}

\maketitle
\thispagestyle{empty}
\pagestyle{empty}

\begin{abstract}

Reinforcement learning (RL) enables agents to take decision based on a
reward function. However, in the process of learning, the choice of
values for learning algorithm parameters can significantly impact the
overall learning process. In this paper, we use a genetic algorithm (GA) to
find the values of parameters used in Deep Deterministic Policy
Gradient (DDPG) combined with Hindsight Experience Replay (HER), to
help speed up the learning agent. We used this method on fetch-reach,
slide, push, pick and place, and door opening in robotic manipulation tasks. Our
experimental evaluation shows that our method leads to better
performance, faster than the original algorithm.
\end{abstract}

\section{INTRODUCTION}
Q-learning methods have been applied on a variety of tasks by autonomous robots \cite{La_TCST2015}, and much research has been done in this field starting many years ago  \cite{watkins1992q}, with some work specific to continuous action spaces \cite{gaskett1999q, doya2000reinforcement, hasselt2007reinforcement, baird1994reinforcement} and others on discrete action spaces 
\cite{wei2017discrete}. Reinforcement Learning (RL) has been applied to locomotion 
\cite{kohl2004policy} \cite{endo2008learning} and also to manipulation
\cite{peters2010relative, kalakrishnan2011learning}.

Much work specific to robotic manipulators also exists 
\cite{deisenroth2011learning, 7864333}. Some of this work used fuzzy wavelet networks \cite{lin2009h}, others used neural networks to accomplish their tasks \cite{miljkovic2013neural}
\cite{duguleana2012obstacle}. Off-policy algorithms such as the Deep Deterministic Policy Gradient algorithm (DDPG)
\cite{lillicrap2015continuous} and Normalized Advantage Function algorithm (NAF) \cite{gu2016continuous} are helpful for real robot systems. A complete review of recent deep reinforcement learning methods for robot manipulation is given in \cite{IRC2019_Nguyen}. We are specifically using DDPG combined with Hindsight Experience Replay (HER) \cite{andrychowicz2017hindsight} for our experiments. Recent work on using experience ranking to improve the learning speed of DDPG + HER was reported in \cite{nguyen2018deep}.

The main contribution of this paper is a demonstration of better final performance at several manipulation tasks using a Genetic Algorithm (GA) to find DDPG and HER parameter values that lead more quickly to better performance at these tasks. Our experiments revealed that learning algorithm parameters are non-linearly related to task performance and learning speed. Rather, success rate can vary significantly based on the values of the parameters used in RL. In the following sections, we describe the manipulation tasks, the DDPG + HER algorithms, and the parameters that affect performance for these algorithms. Initial experimental results showing performance and speed gains when using a GA to search for good parameter values then provide evidence that GAs find good parameter values leading to better task performance, faster.

The paper is organized as follows: In Section 2, we present related work. Section 3 describes the DDPG + HER algorithms. In Section 4, we describe the GA being used to find the values of parameters. Section 5 then describes our learning tasks and experiments and our experimental results. The last section provides conclusions and possible future research.

\section{RELATED WORK}
RL has been widely used in training/teaching both a single robot \cite{pham2018, Pham_SSRR2018} and a multi-robot system \cite{La_CYBER2013,  La_SMCA2015, pham2018cooperative, Dang_MFI2016, rahimi2018}. Previous work has also been done on both model-based and model-free learning algorithms. Applying model-based learning algorithms to real world scenarios, rely significantly on a model-based teacher to train deep network policies. 

Similarly, there is also much work in GA's \cite{davis1991handbook}
\cite{deb2002fast} and the GA operators of crossover
and mutation \cite{poon1995genetic}, applied to a variety of
problem. GA has been specifically applied to variety of RL problems \cite{liu2009study,moriarty1999evolutionary,mikami1994genetic,poon1995genetic}.

In this paper, we use model-free RL with
continuous action spaces and deep neural network. Our work is built on
existing work using the same techniques applied to robotic manipulator
\cite{lillicrap2015continuous} \cite{andrychowicz2017hindsight}.
Specifically, we use a GA to search for good DDPG + HER algorithm parameters and compare it with original values of parameters \cite{baselines}, and hence the success rates. DDPG + HER, a RL algorithm using deep neural networks in continuous action spaces has been successfully used for robotic manipulation tasks, and our GA improves on this work by finding learning algorithm parameters that needs fewer epochs (one epoch is a single pass through full training set) to learn better task performance.

\section{BACKGROUND}

\subsection{Reinforcement Learning}
Consider a standard RL setup consisting of a learning agent, which interacts with an environment. An environment can be described by a set of variables where $S$ is the set of states, $A$ is the set of actions, $p(s_0)$ is a distribution of initial states, $r: S \times A \xrightarrow{} R$, $p(s_{t+1}|s_t,a_t)$ are transition probabilities and $\gamma \in [0,1]$ is a discount factor.  

A deterministic policy maps from states to actions: $\pi : S \xrightarrow{} A$. The beginning of every episode is marked by sampling an initial state $s_0$. For each timestep $t$, the agent performs an action based on the current state: $a_t = \pi(s_t)$. The performed action gets a reward $r_t = r(s_t, a_t)$, and the distribution $p(.|s_t,a_t)$ helps to sample the environment’s new state. The total return is: $R_t = \sum_{i=T}^\infty \gamma^{i-t}r_i$ . The agent’s goal is to try to maximize its expected return $E[R_t|s_t,a_t]$ and an optimal policy denoted by $\pi^*$ can be defined as any policy $\pi^*$ , such that $Q^{\pi^*} (s,a) \geq Q^\pi (s,a)$ for every $s \in S, a \in A$ and any policy $\pi$. The optimal policy, which has the same Q-function, is called an optimal Q-function, $Q^*$ , which satisfies the \textit{Bellman} equation:   
\begin{gather} 
    Q^*(s,a) = E_{s'~p(.|s,a))} [r(s,a) + \gamma \underset{a'\in A}{max} Q^*(s',a'))]. \label{equation}
\end{gather}

\subsection{Deep Q-Networks(DQN)}

A \textit{Deep Q-Networks (DQN)} \cite{wang2015dueling} is defined as a model free reinforcement learner, designed for discrete action spaces. In a DQN, a neural network $Q$ is maintained, which approximates $Q^*$. $\pi_Q(s) = argmax_{a\in A}Q(s,a)$ denotes a greedy policy w.r.t. $Q$. A - greedy policy takes a random action with probability $\epsilon$ and action $\pi_Q(s)$ with probability $1 - \epsilon$ .   
 
Episodes are generated during training using a $\epsilon$-greedy policy. A \textit{Replay buffer} stores transition tuples $(s_t,a_t,r_t,s_{t+1})$ experienced during training. The neural network training is interlaced by generation of new episodes. A Loss $\mathcal{L}$ defined by $\mathcal{L}=E(Q(s_t,a_t)-y_t)^2$ where $y_t=r_t+\gamma max_{a'\in A}Q(s_{t+1},a')$ and tuples $(s_t,a_t,r_t,s_{t+1})$ are being sampled from the replay buffer. 

The \textit{target network} changes at a slower pace than the main network, which is used to measure targets $y_t$. The weights of the target networks can be set to the current weights of the main network \cite{wang2015dueling}. Polyak-averaged parameters \cite{polyak1992acceleration} can also be used.  
 
\subsection{Deep Deterministic Policy Gradients (DDPG)}

In \textit{Deep Deterministic Policy Gradients (DDPG)}, there are two neural networks: an Actor and a Critic. The actor neural network is a target policy $\pi :S\xrightarrow{}A$, and critic neural network is an action-value function approximator $Q:S\times A\xrightarrow{}R$. The critic network $Q(s,a|\theta^Q)$ and actor network $\mu(s|\theta^\mu)$ are randomly initialized with weights $\theta^Q$ and $\theta^\mu$. 

A behavioral policy is used to generate episodes, which is a noisy variant of target policy, $\pi_b(s)=\pi(s) + \mathcal{N}(0,1)$. The training of a critic neural network is done like the Q-function in DQN but where the target $y_t$ is computed as $y_t=r_t+\gamma Q(s_{t+1},\pi(s_{t+1}))$, where $\gamma$ is the discounting factor. The loss $\mathcal{L}_a=-E_aQ(s,\pi(s))$ is used to train the actor network.

\subsection{Hindsight Experience Replay (HER)}

Hindsight Experience Reply (HER) tries to mimic human behavior to learn from failures. The agent learns from all episodes, even when it does not reach the original goal. Whatever state the agent reaches, HER considers that as the modified goal. Standard experience replay only stores the transition $(s_t||g,a_t,r_t,s_{t+1}||g)$ with original goal $g$. HER tends to store the transition $(s_t||g',a_t,r'_t,s_{t+1}||g')$ to modified goal $g'$ as well. HER does great with extremely sparse rewards and is also significantly better for sparse rewards than shaped ones.    

\subsection{Genetic Algorithm (GA)}

\textit{Genetic Algorithms (GAs)} \cite{davis1991handbook, holland1992genetic, goldberg1988genetic} were designed to search poorly-understood spaces, where exhaustive search may not be feasible, and where other search approaches perform poorly. When used as function optimizers, GAs try to maximize a fitness tied to the optimization objective. Evolutionary computing algorithms in general and GAs specifically have had much empirical success on a variety of difficult design and optimization problems. They start with a randomly initialized population of candidate solution typically encoded in a string (chromosome). A selection operator focuses search on promising areas of the search space while crossover and mutation operators generate new candidate solutions. We explain our specific GA in the next section.

\begin{figure}[h!]
\centering
\hspace*{-1cm}
  \begin{subfigure}[b]{0.8\linewidth}
    \includegraphics[width=8cm,height=6cm]{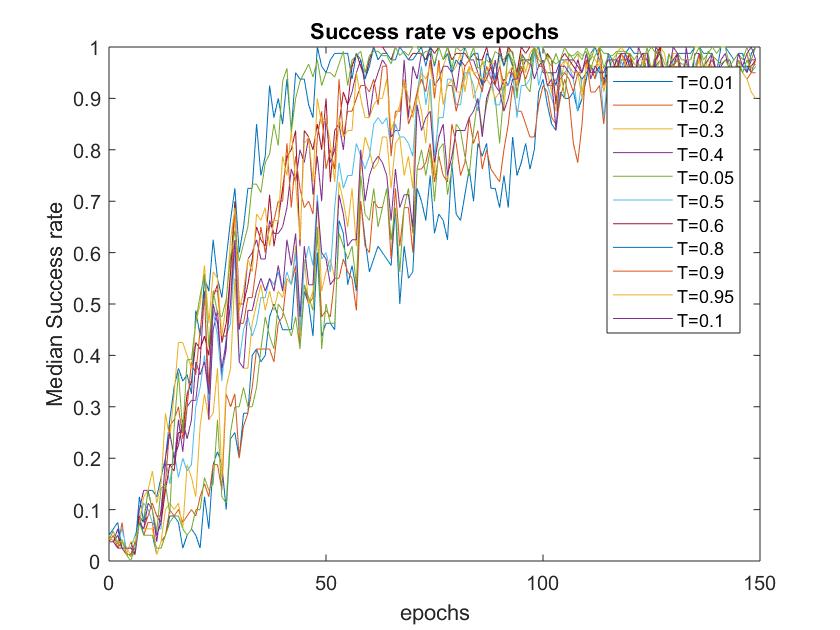}
  \end{subfigure}
  \begin{subfigure}[b]{0.9\linewidth}
    \includegraphics[width=8cm,height=6cm]{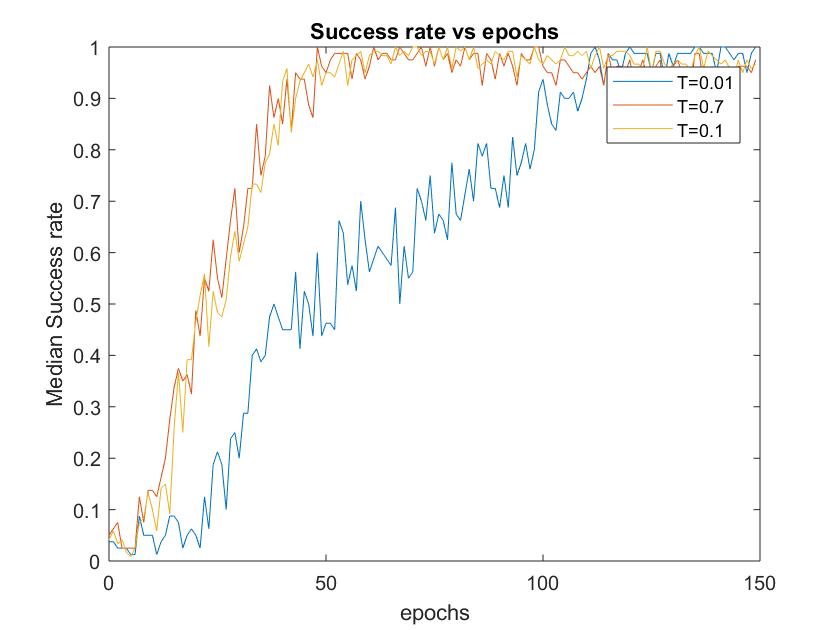}
  \end{subfigure}
  \begin{subfigure}[b]{0.9\linewidth}
    \includegraphics[width=8cm,height=6cm]{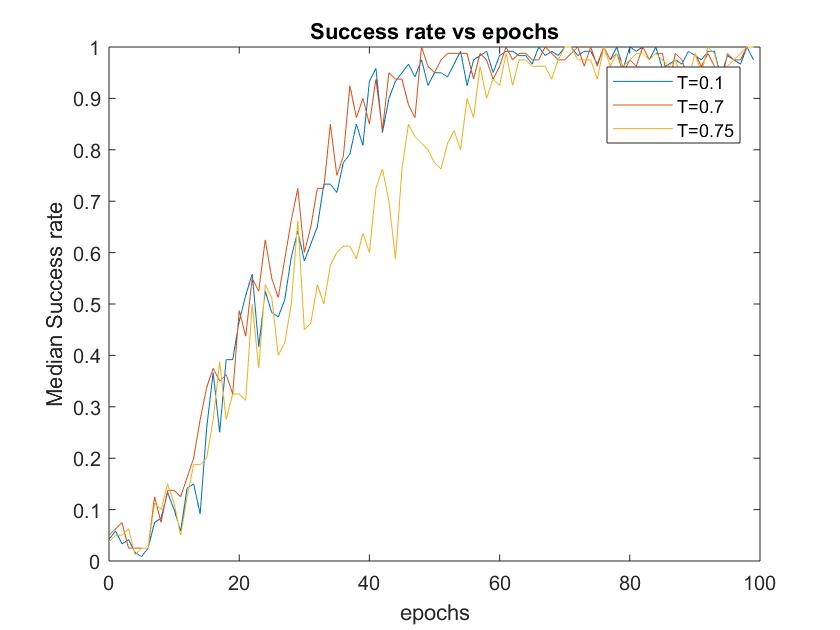}
  \end{subfigure}
  \caption{Success rate vs. epochs for various $\tau$ for \textit{FetchPick\&Place-v1} task.}
  \label{fig:TargetReaching}
\end{figure}

\begin{figure}[h!]
\centering
\hspace*{-1cm}
  \begin{subfigure}[b]{0.8\linewidth}
   \centering
    \includegraphics[width=8.6cm,height=6.5cm]{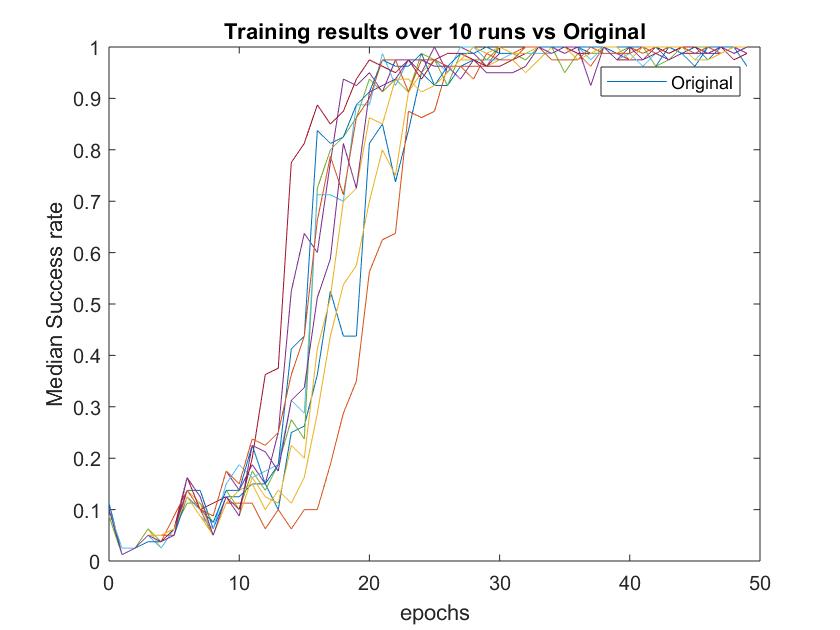}
    \caption{Optimal Parameters over 10 runs, vs. Original}\label{fig:1a}
  \end{subfigure}
  \begin{subfigure}[b]{0.9\linewidth}
  \centering
    \includegraphics[width=8.6cm,height=6.5cm]{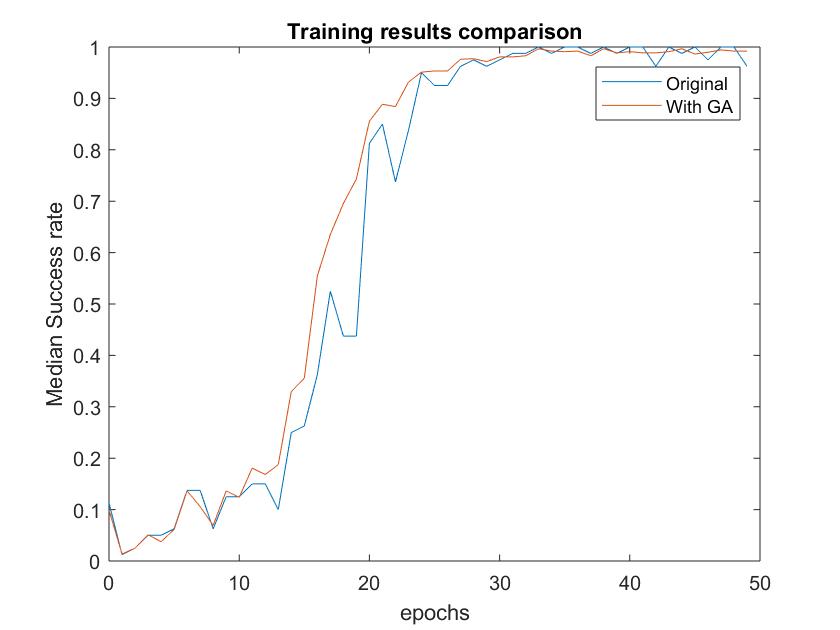}
    \caption{Optimal Parameters averaged over 10 runs, vs. Original}\label{fig:1a}
  \end{subfigure}
  \caption{Success rate vs. epochs for \textit{FetchPush-v1} task when $\tau$ and $\gamma$ are found using the GA.}
  \label{fig:compareFetchPush}
\end{figure}

\begin{figure}[h!]
\centering
\hspace*{-1cm}
  \begin{subfigure}[b]{0.8\linewidth}
   \centering
    \includegraphics[width=8.6cm,height=6.5cm]{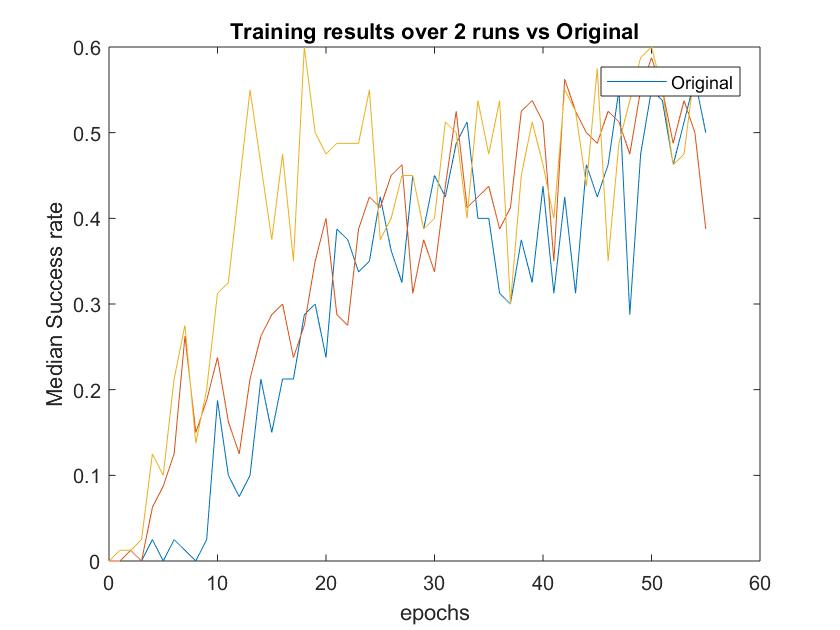}
    \caption{Optimal Parameters over 2 runs, vs. Original}\label{fig:1a}
  \end{subfigure}
  \begin{subfigure}[b]{0.9\linewidth}
  \centering
    \includegraphics[width=8.6cm,height=6.5cm]{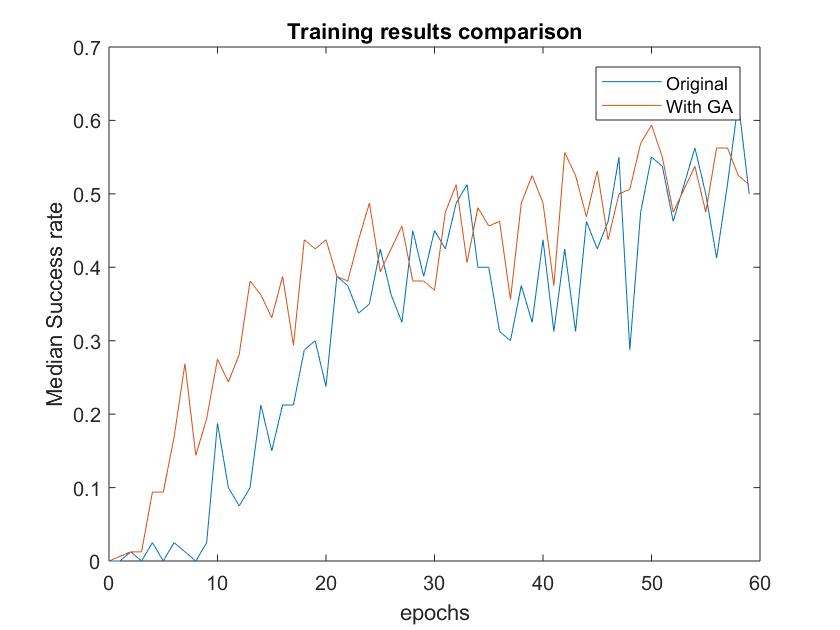}
    \caption{Optimal Parameters averaged over 2 runs, vs. Original}\label{fig:1a}
  \end{subfigure}
  \caption{Success rate vs. epochs for \textit{FetchSlide-v1} task when $\tau$ and $\gamma$ are found using the GA.}
  \label{fig:compareFetchSlide}
\end{figure}

\section{DDPG + HER and GA}
In this section, we present the primary contribution of our paper: The genetic algorithm searches through the space of parameter values used in DDPG + HER for values  that maximize task performance and minimize the number of training epochs. We target the following parameters: discounting factor $\gamma$; polyak-averaging coefficient $\tau$ \cite{polyak1992acceleration}; learning rate for critic network $\alpha_{critic}$; learning rate for actor network $\alpha_{actor}$; percent of times a random action is taken $\epsilon$; and standard deviation of Gaussian noise added to not completely random actions as a percentage of maximum absolute value of actions on different coordinates $\eta$. The range of all the parameters is 0-1, which can be justified using the equations following in this section.  

Our experiments show that adjusting the values of parameters did not increase or decrease the agent’s learning in a linear or easily discernible pattern. So, a simple hill climber will probably not do well in finding optimized parameters. Since GAs were designed for such poorly understood problems, we use our GA to optimize these parameter values.  

Specifically, we use $\tau$ , the polyak-averaging coefficient to show the performance non-linearity for values of $\tau$ . $\tau$ is used in the algorithm as show in Equation (\ref{tau}):  

\begin{gather} 
    \theta^{Q'} \xleftarrow{} \tau\theta^Q+(1-\tau)\theta^{Q'},
    \nonumber\\
    \theta^{\mu'} \xleftarrow{} \tau\theta^\mu+(1-\tau)\theta^{\mu'}.
    \label{tau}
\end{gather}

\begin{algorithm}
\caption{DDPG + HER and GA}\label{euclid}
\begin{algorithmic}[1]
\State Choose population of $n$ chromosomes
\State Set the values of parameters into the chromosome
\State Run the DDPG +  HER to get number of epochs for which the algorithm first reaches success rate $\geq 0.85$
\For{all chromosome values} 
    \State Initialize DDPG
    \State Initialize replay buffer $R \gets \phi$
    \For{episode=1, M}
        \State Sample a goal $g$ and initial state $s_0$
        \For{t=0, T-1}
            \State Sample an action $a_t$ using DDPG behavioral policy
            \State Execute the action $a_t$ and observe a new state $s_{t+1}$
        \EndFor
        \For{t=0, T-1}
            \State $r_t:=r(s_t,a_t,g)$
            \State Store the transition $(s_t||g,a_t,r_t,s_{t+1}||g)$ in $R$
            \State Sample a set of additional goals for replay $G:=S$(\textbf{current episode})
            \For{$g'\in G$}
                \State $r':=r(s_t,a_t,g')$
                \State Store the transition $(s_t||g',a_t,r',s_{t+1}||g')$ in $R$ 
            \EndFor
        \EndFor
        \For{t=1,N}
            \State Sample a minibatch $B$ from the replay buffer $R$
            \State Perform one step of optimization using $A$ and minibatch $B$
        \EndFor
    \EndFor
    \State \textbf{return} $1/epochs$
\EndFor
\State Perform Uniform Crossover
\State Perform Flip Mutation at rate 0.1
\State Repeat for required number of generations to find optimal solution
\end{algorithmic}
\end{algorithm}

Equation (\ref{target y}) shows how $\gamma$ is used in the DDPG + HER algorithm, while Equation (\ref{q learning}) describes the Q-Learning update. α denotes the learning rate. Networks are trained based on this update equation.
\begin{gather} 
    y_i = r_i + \gamma Q'(s_{i+1},\mu'(s_{t+1}|\theta^{\mu'})|\theta^{Q'}), \label{target y}
\end{gather}
\begin{gather} 
    Q(s_t,a_t) \xleftarrow{} Q(s_t,a_t) + \alpha[r_{t+1} + \gamma Q(s_{t+1},a_{t+1}) \nonumber\\ - Q(s_t,a_t)].
    \label{q learning}
\end{gather}

Since we have two kinds of networks, we will need two learning rates, one for the actor network ($\alpha_{actor}$), another for the critic network ($\alpha_{critic}$). Equation (\ref{action}) explains the use of percent of times that a random action is taken, $\epsilon$.

\begin{gather} 
a_t = 
\begin{cases}
     a^*_t \qquad \qquad \qquad \; \; \; \; \; \, with\;probability\;1 - \epsilon, \\
     random\;action \qquad  with\;probability\;\epsilon.
\end{cases}
\label{action}
\end{gather}

Figure \ref{fig:TargetReaching} shows that when the value of $\tau$ is modified, there is a change in the agent’s learning, further emphasizing the need to use a GA. The original (untuned) value of $\tau$ in DDPG was set to 0.95, and we are using 4 CPUs. All the values of $\tau$ are considered up to two decimal places, in order to see the change in success rate with change in value of the parameter. From the plots, we can clearly tell that there is a great scope of improvement from the original success rate. 

Algorithm 1 explains the integration of DDPG + HER  with a GA, which uses a population size of 30 over 30 generations. We are using \textit{ranking selection} 
\cite{goldberg1991comparative} to select parents. The parents are probabilistically based on rank, which is in turn decided based on the relative fitness (performance). Children are then generated using \textit{uniform crossover}
\cite{syswerda1989uniform}. We are also using \textit{flip mutation}
\cite{goldberg1988genetic} with probability of mutation to be 0.1. We use a binary chromosome to encode each parameter and concatenate the bits to form a chromosome for the GA. The six parameters are arranged in the order: polyak-averaging coefficient; discounting factor; learning rate for critic network; learning rate for actor network; percent of times a random action is taken and standard deviation of Gaussian noise added to not completely random actions as a percentage of maximum absolute value of actions on different coordinates. Since each parameter requires 11 bits to be represented to three decimal places, we need 66 bits for 6 parameters. These string chromosomes then enable domain independent crossover and mutation string operators to generate new parameter values. We consider parameter values up to three decimal places, because small changes in values of parameters causes considerable change in success rate. For example, a step size of 0.001 is considered as the best fit for our problem. 

The fitness for each chromosome (set of parameter values) is defined by the inverse of number of epochs it takes for the learning agent to reach close to maximum success rate ($\geq0.85$) for the very first time. Fitness is the inverse of number of epochs because GA always maximizes the objective function and this converts our minimization of number of epochs to a maximization problem. Since each fitness evaluation takes significant time an exhaustive search of the $2^{66}$ size search space is not possible and we thus use GA search.


\begin{figure}
\centering
\begin{multicols}{2}
    \includegraphics[width=4cm,height=3.1cm]{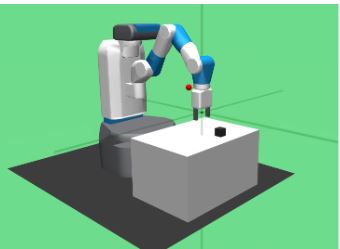}
    \vspace{0.1cm}
    \subcaption{FetchPick\&Place environment}
    \vspace{0.1cm}
    \includegraphics[width=4cm,height=3.1cm]{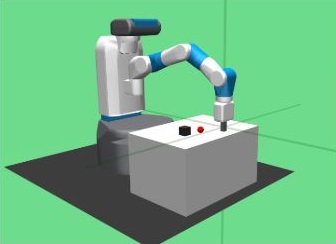}
    \vspace{0.1cm}
    \subcaption{FetchPush environment}
    \vspace{0.1cm}
    \includegraphics[width=4cm,height=3.1cm]{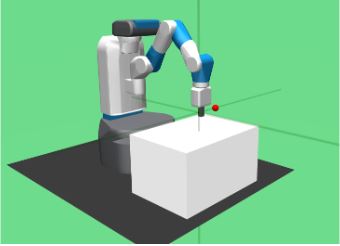}
    \subcaption{FetchReach environment}
    \vspace{0.1cm}
    \includegraphics[width=4cm,height=3cm]{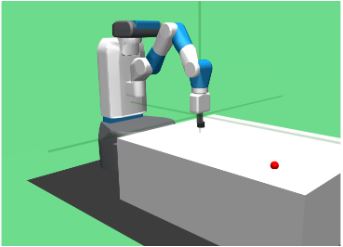}
    \subcaption{FetchSlide environment}
    \vspace{0.1cm}
    \includegraphics[width=4cm,height=3.3cm]{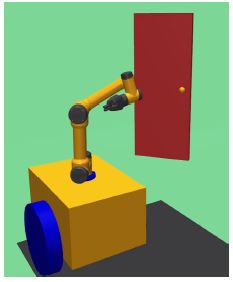}
    \subcaption{Door Opening environment}
    \vspace{0.1cm}
    \includegraphics[width=4.6cm,height=3.4cm]{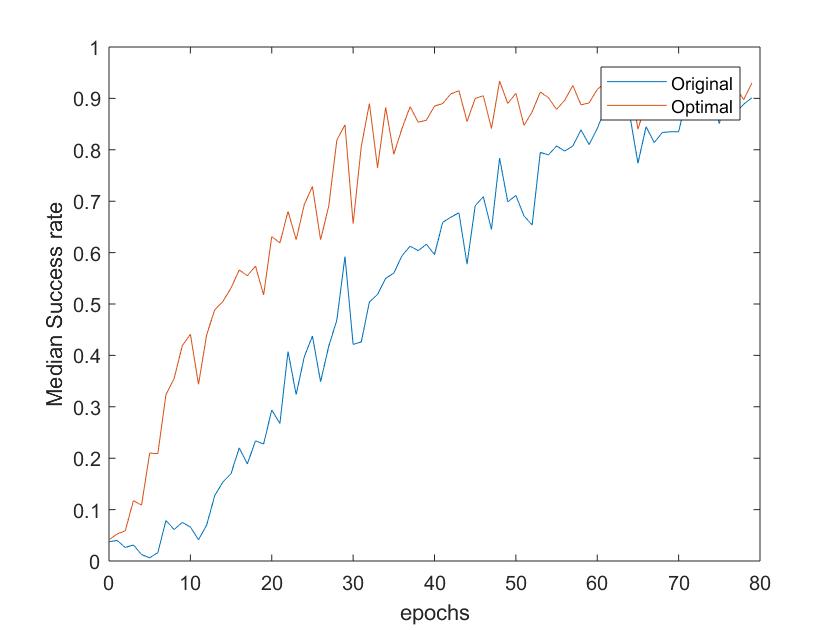}
    \subcaption{FetchPick\&Place plot}
    \vspace{0.1cm}
    \includegraphics[width=4.6cm,height=3.4cm]{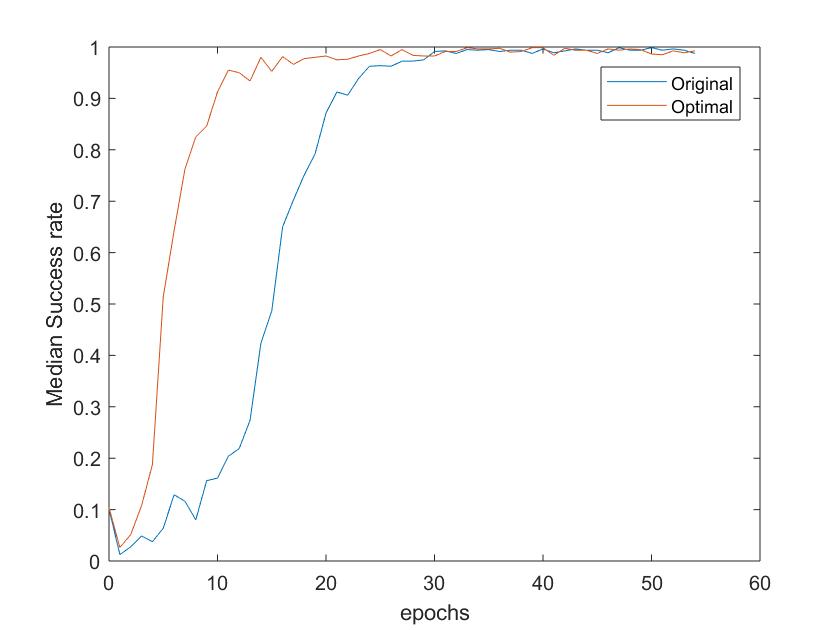}
    \subcaption{FetchPush plot}
    \includegraphics[width=4.6cm,height=3.2cm]{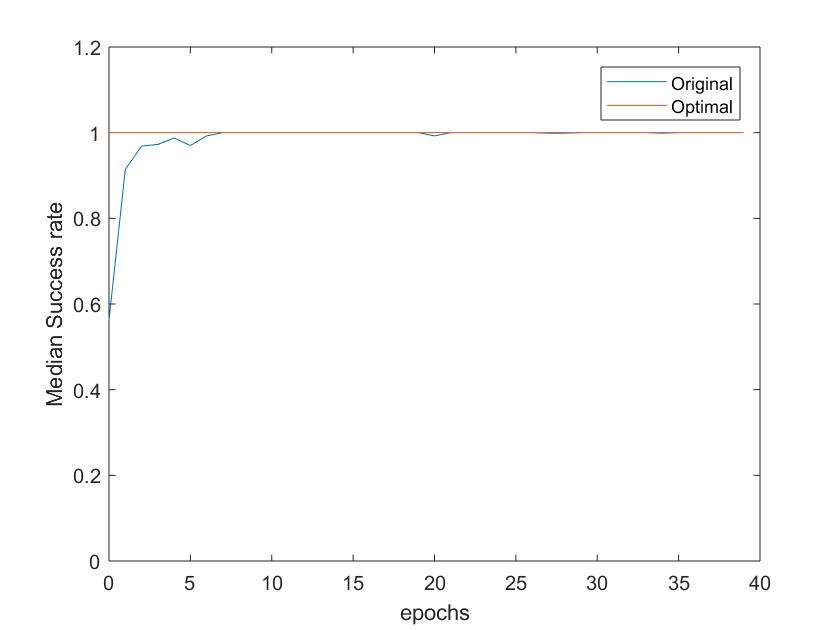}
    \subcaption{FetchReach plot}
    \vspace{0.1cm}
    \includegraphics[width=4.6cm,height=3.2cm]{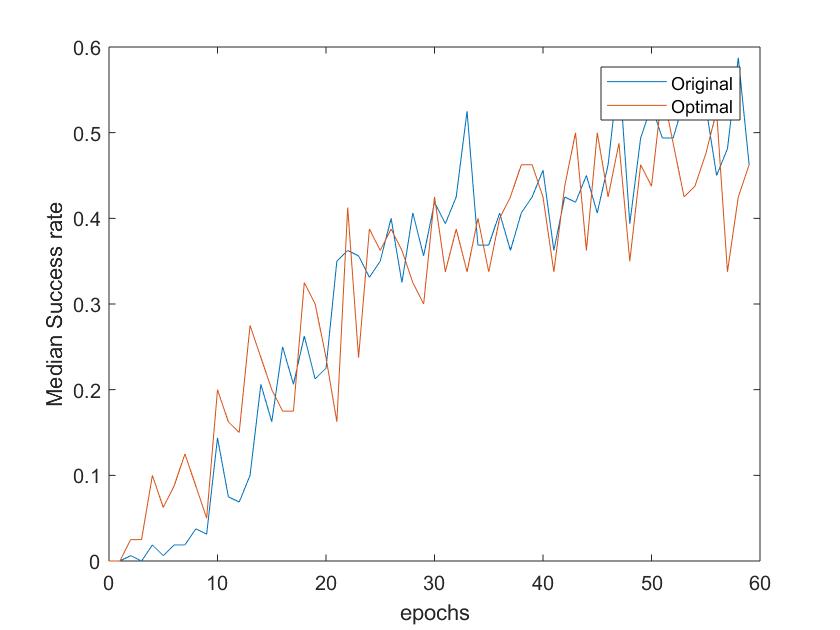}
    \subcaption{FetchSlide plot}
    \includegraphics[width=4.6cm,height=3.2cm]{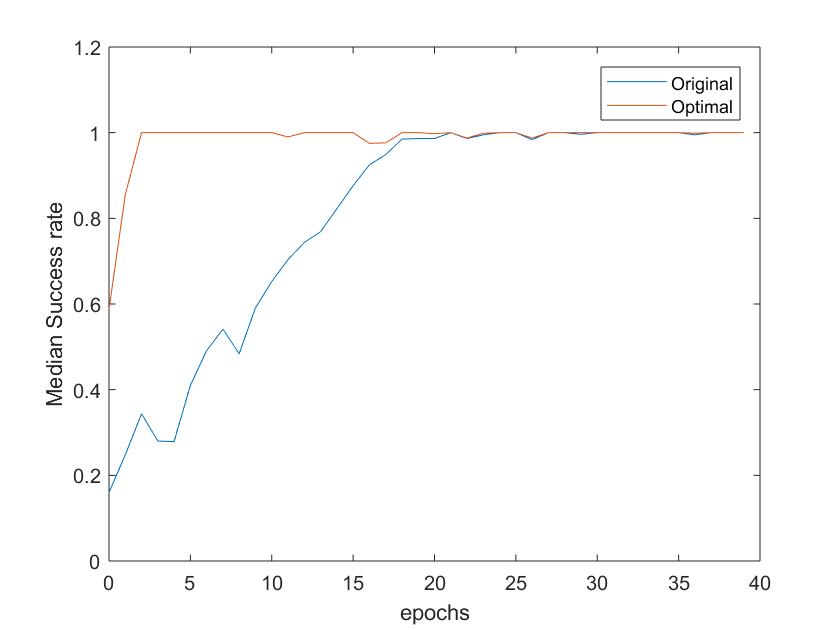}
    \subcaption{DoorOpening plot}
\end{multicols}
\caption{Environments and the corresponding Original vs Optimal plots, when all the 6 parameters are found by GA }
\label{finalPlots}
\end{figure}

\section{EXPERIMENT and RESULTS}
Figure \ref{finalPlots}, shows the environments used to test robot learning on five different tasks: \textit{FetchPick\&Place-v1}, \textit{FetchPush-v1},   \textit{FetchReach-v1},  \textit{FetchSlide-v1}, and \textit{DoorOpening} . We ran the GA separately on these environments to check the effectiveness of our algorithm and compared performance with the original values of the parameters. Figure \ref{fig:compareFetchPush} (a) shows the result of our experiment with \textit{FetchPush-v1}, while Figure \ref{fig:compareFetchSlide} (a) shows the results with \textit{FetchSlide-v1}. We let the system run with GA to find the optimal parameters τ and γ. Since the GA is probabilistic, we show results from 10 runs of the GA and the results show that the optimized parameters found by the GA can lead to better performance. The learning agent can run faster, and can reach the maximum success rate, faster. In Figure \ref{fig:compareFetchPush} (b), we show one learning run for the original parameter set and the average learning over these 10 different runs of the GA.

\begin{table}[h!]
\centering
\begin{tabular}{||c c c||} 
 \hline
 Parameters & Original & Optimal \\ [0.5ex] 
 \hline\hline
 $\gamma$ & 0.98 & 0.88\\ 
 $\tau$ & 0.95 & 0.184\\
 $\alpha_{actor}$ & 0.001 & 0.001 \\
 $\alpha_{critic}$ & 0.001 & 0.001 \\
 $\epsilon$ & 0.3 & 0.055\\ 
 $\eta$ & 0.2 & 0.774 \\[1ex] 
 \hline
\end{tabular}
\caption{Original vs Optimal values of parameters}
\label{table:1}
\end{table}

Figure \ref{fig:compareFetchSlide} (b) compares one run for original with averaged 2 runs for optimizing parameters $\tau$ and $\gamma$. For this task, we have run it for only 2 runs because these tasks can take a few hours for one run. The results shown in Figures \ref{fig:compareFetchPush} and \ref{fig:compareFetchSlide} show changes when only two parameters are being optimized as we tested and debugged the genetic algorithm be we can see the possibility for performance improvement. Our results from optimizing all five parameters justify this optimism and are described next. 

The GA was then run to optimize all parameters and these results were plotted in Figure \ref{finalPlots} for all the tasks. Table \ref{table:1} compares the GA found parameters with the original parameters used in the RL algorithm. Though the learning rates $\alpha_{actor}$ and $\alpha_{critic}$ are same as their original values, the other four parameters have different values than original. The plots in the figure \ref{finalPlots} shows that the GA found parameters outperformed the original parameters, indicating that the learning agent was able to learn faster. All the plots in this figure are averaged over 10 runs.  

\section{DISCUSSION and FUTURE WORK}

In this paper, we showed initial results that demonstrated that a genetic algorithm can tune reinforcement learning algorithm parameters to achieve better performance, faster at six manipulation tasks. We discussed existing work in reinforcement learning in robotics, presented an algorithm, which integrates DDPG + HER with GA to optimize the number of epochs required to achieve maximal performance, and explained why a GA might be suitable for such optimization. Initial results bore out the assumption that GAs are a good fit for such parameter optimization and our results on the six manipulation tasks show that the GA can find parameter values that lead to faster learning and better (or equal) performance at our chosen tasks. We thus provide further evidence that heuristic search as performed by genetic and other similar evolutionary computing algorithms are a viable computational tool for optimizing reinforcement learning performance in multiple domains. 


\section*{APPENDIX}

We have the code for this paper on github: \textit{https://github.com/aralab-unr/ReinforcementLearningWithGA}. The parameters used in this paper can be found in \textit{baselines.her.experiment.config} module. The parameters are: discounting factor; polyak-averaging coefficient; learning rate for critic network; learning rate for actor network; percent of times a random action is taken; and standard deviation of Gaussian noise added to not completely random actions as a percentage of maximum absolute value of actions on different coordinates, corresponds to $gamma$; $polyak$; $Q\_lr$; $pi\_lr$; $random\_eps$, $noise\_eps$, respectively in the code.

\bibliography{root.bib}
\bibliographystyle{IEEEtran}

\end{document}